\documentclass[9pt]{article}
\usepackage[colorinlistoftodos]{todonotes}
\usepackage{siunitx}
\usepackage{booktabs} 
\usepackage{graphicx}
\usepackage{placeins}   
\usepackage{stfloats}
\usepackage{spconf,amsmath,graphicx,hyperref}
\usepackage{subcaption}
\usepackage{amssymb}
\usepackage[toc,page]{appendix}

\title{MATTER: Multiscale Attention for Registration Error Regression\vspace{-.3cm}}
\name{Shipeng Liu, Ziliang Xiong, Khac-Hoang Ngo, Per-Erik Forssén 
\thanks{This work was funded by Swedish national strategic research environment ELLIIT, including grant C08. Training and inference used resources provided by the National Academic Infrastructure for Supercomputing in Sweden (NAISS), partially funded by the Swedish Research Council through grant agreement no. 2022-06725. This project is also supported by the Wallenberg AI, Autonomous Systems, and Software Program (WASP) funded by the Knut and Alice Wallenberg Foundation.}
\vspace{-.4cm}
}
\address{Department of Electrical Engineering, Linköping University, Linköping, Sweden
\vspace{-.5cm}}
\begin{document}
%
\maketitle
\begin{abstract}
Point cloud registration (PCR) is crucial for many downstream tasks, such as simultaneous localization and mapping (SLAM) and object tracking. This makes detecting and quantifying registration misalignment, i.e.,~{\it PCR quality validation}, an important task.
All existing methods treat validation as a classification task, aiming to assign the PCR quality to a few classes. 
In this work, we instead use regression for PCR validation, allowing for a more fine-grained quantification of the registration quality.
We also extend previously used misalignment-related features by using multiscale extraction and attention-based aggregation.
This leads to accurate and robust registration error estimation on diverse datasets, especially for point clouds with heterogeneous spatial densities.
Furthermore, when used to guide a mapping downstream task, our method significantly improves the mapping quality for a given amount of re-registered frames, compared to the state-of-the-art classification-based method.

\end{abstract}

\vspace{-3mm}

\section{Introduction}
\label{sec:intro}

Point cloud registration (PCR) is a core problem in 3D computer vision, aiming to estimate the rigid transform between two overlapping point clouds. It underpins numerous downstream applications, including simultaneous localization and mapping (SLAM), 3D reconstruction, and robotic navigation. 
Despite recent advances in PCR, registration error is still significant in challenging scenarios, e.g., when optimization of the loss function 
gets trapped in local minima  \cite{access_loss}, under motion distortion \cite{loam_ji} or measurement noise, or when the problem is geometrically under-constrained~\cite{pred_nobili, incor_henrik}.
Errors in PCR can propagate to downstream modules, resulting in map distortion and navigation failure. This motivates the need for registration misalignment detection, i.e., predicting, after registration, whether the registered point clouds deviate from the well-aligned 
ground truth. 
This prediction can then allow downstream tasks to take corrective action.

\begin{figure}[h]
  \centering
  \includegraphics[clip,trim={0 0 0 13mm},width=0.9\linewidth]{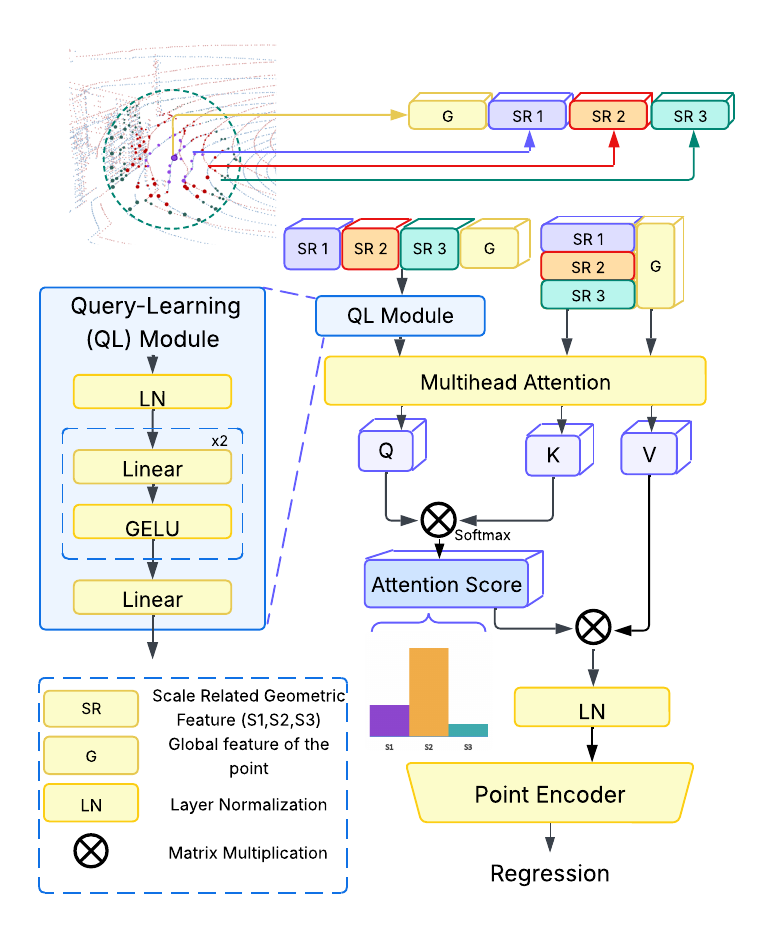}
  \vspace{-.6cm}
  \caption{Overview of the proposed multiscale attention mechanism for geometric features in registered point clouds.}
  \vspace{-6mm}
  \label{fig:method}
\end{figure}

The objective functions used in existing registration methods, such as closest-point residuals~\cite{132043}, score functions~\cite{liao2020point}, and entropy gain~\cite{tustison2010point}, are not evaluated against the ground truth pose and are sensitive to local minima \cite{access_loss}, noise, and scene geometry, their final values do not reliably reflect the true alignment quality.
Recent works~\cite{almqvist2018learning, bogoslavskyi2017analyzing, camous2022deep, dillen2025fact, yin2019failure, adolfsson2021coral} proposed learning-based approaches for point cloud misalignment classification (PCMC), where the registration result is categorized into discrete quality levels. Specifically, in \cite{yin2019failure, almqvist2018learning}, logistic regression is used with manually designed features for binary classification.
 CorAl, introduced in~\cite{adolfsson2021coral}, fits a 3D Gaussian distribution to the neighborhood of each point, and uses the differential entropy of this distribution as the feature. FACT~\cite{dillen2025fact}\textemdash the state-of-the-art method\textemdash extends binary classification to multiple classes, which allows evaluation at a finer level. 
However, all these works are limited to classification, i.e., coarsely categorizing the misalignment into a few levels, without providing continuous predictions of the actual alignment error. A more comprehensive review of related work is provided in Appendix~\ref{appendix:relatedwork}.

We here recast PCMC as a continuous prediction problem, regressing a scalar alignment error from multiscale geometric features fused by attention. Our contributions are threefold. First, we extend PCMC to point cloud misalignment regression~(PCMR) for fine-grained assessment. Second, we introduce MATTER (Multiscale ATTention for registration Error Regression), which improves robustness by extracting features at multiple geometric scales and fusing them using attention. The MATTER architecture is shown in Fig.~\ref{fig:method}, and detailed in Section~\ref{sec:metho}.
Third, we adapt prior PCMC methods to regression and demonstrate through experiments on three datasets, that MATTER consistently outperforms them. We also apply MATTER to a mapping downstream task, to detect misaligned frames that need correction. 
Under the same re-registration budget, MATTER achieves a better map quality than the state-of-the-art methods.


\vspace{-.2cm}
\section{Methods}
\label{sec:metho}
\vspace{-.1cm}
In this section, we first define the PCMR task, introduce the overall architecture of MATTER and describe the proposed multiscale attention depicted in Fig~\ref{fig:method}. 

\vspace{-3mm}

\subsection{Task Formulation} 
Consider a pair of source and reference point clouds $(P^\mathrm{S}, P^\mathrm{R})$, 
which are sets of points in $\mathbb{R}^3$ that have been registered by an estimated rigid transform $\hat{T}\in\mathbb{SE}(3)$. We define their alignment error
as
the point-wise mean distance between the source point cloud aligned by~$\hat{T}$ and the same source aligned by the ground-truth transform, $T^\ast$~\cite{dillen2025fact},
\vspace{-2.0mm}
\begin{equation}
E_{\text{align}} \;=\; 
\,\frac{1}{|P^\mathrm{S}|}\sum_{p \in P^\mathrm{S}} \big\|\,\hat{T}(p)\,-\,T^*(p)\,\big\|_2\;.
\label{eq:align_error}\vspace{-2mm}
\end{equation}
We aim to regress this error to obtain a fine-grained estimate of the misalignment. 

\vspace{-3mm}

\subsection{Model Architecture}
We adapt the FACT architecture~\cite{dillen2025fact} to the PCMR task by using multiscale attention to fuse alignment-related features, see Fig.~\ref{fig:method}.
Following \cite{dillen2025fact}, we extract global and local neighbourhood features at multiple scales.
These features are then concatenated and cross-attended to extract scale-wise attention weights.
The weights are then used to reweight the multiscale features before passing to a point transformer and regression head for the final prediction.
Training uses an $l_2$ loss.

\vspace{-3mm}

\subsection{Feature Extraction}
We first transform both $P^\mathrm{S}$ and $P^\mathrm{R}$ point clouds into a common frame and apply farthest point sampling~\cite{fps} 
to select the same number of anchor points from each of them. 
In the rest of the section, we focus on a generic anchor point. Let $\mathcal{N}_{P^\mathrm{S}}$ and $\mathcal{N}_{P^\mathrm{R}}$ denote the points from the source and reference point clouds, respectively, within the neighborhood of this anchor point.
We use $\mathcal{N}_\mathrm{sep}$ to refer to $\mathcal{N}_{P^\mathrm{S}}$ if the anchor point belong to $P^\mathrm{S}$, and to $\mathcal{N}_{P^\mathrm{R}}$ if it belongs to $P^\mathrm{R}$.
Define the joint neighborhood $\mathcal{N}_{\mathrm{joint}}=\mathcal{N}_{P^\mathrm{S}}\cup\mathcal{N}_{P^\mathrm{R}}$.
Following \cite{dillen2025fact}, we extract local features around each anchor, namely the separate and joint differential entropy~\cite{adolfsson2021coral} $H(\mathcal{N}{\mathrm{sep}}), H(\mathcal{N}{\mathrm{joint}})$, and the Sinkhorn divergence $D_\lambda$.
They measure the discrepancy between the distributions of points in the aligned source with respect to reference neighborhoods. Additional details on the feature computation are provided in Appendix~\ref{appendix:features}.
Unreliable neighborhoods are handled explicitly: if one point cloud contributes no points, the differential entropy and Sinkhorn divergence are set to a bounded default.
The next local features are coverage ratios $\rho_{\mathrm{joint}}=\big|\mathcal{N}_{\mathrm{joint}}\big|/(\big|P^\mathrm{S}\big|+\big|P^\mathrm{R}\big|)$ and
$\rho_{\mathrm{sep}}=\big|\mathcal{N}_{\mathrm{sep}}\big|/\big|P_{\mathrm{sep}}\big|$, where $\big|P_{\mathrm{sep}}\big|$ is given by $ \big|P^\mathrm{{S}}\big|$ or $\big|P^\mathrm{{R}}\big|$ if the anchor point belongs to $P^\mathrm{{S}}$ or $P^\mathrm{{R}}$, respectively.

Three global point features are computed on the whole point cloud: (1) a co-visibility score $c\!\in[0,1]$ computed with a visibility operator~\cite{visibility}, (2) the distance $d$ from the point to the LiDAR sensor, and (3) a binary source flag $b \in \{0,1\}$ designating which point cloud the point originates from. These features serve to down-weight non-covisible or distant regions, while up-weighting well-observed ones.

\vspace{-3.3mm}

\subsection{Multiscale Cross-Attention}


\noindent \textbf{Multiscale Features.}
In FACT~\cite{dillen2025fact}, the radius of $\mathcal{N}_{\mathrm{joint}}$ is chosen as a single value, adaptively adjusted according to $d$. There is a tradeoff between using small neighborhoods and large ones.
A small radius may fail when the initial alignment is poor, since the corresponding structure can fall outside the neighborhoods;
in contrast, a large neighborhood contains non-overlapping regions, which makes the entropy and Sinkhorn divergence less reliable for estimating registration errors.
(We will illustrate this in Fig.~\ref{fig:pc_all}.) Adapting a single radius according to $d$, as in FACT~\cite{dillen2025fact}, does not capture well this tradeoff. Moreover, a larger neighborhood increases distributional heterogeneity and thus deviates from the approximately isotropic Gaussian assumption. This undermines the validity of the closed-form differential-entropy estimate.

To resolve these failure modes, we compute features at $S$ different radii for every anchor point, and fuse these features with attention.
For each scale $s \in \{1, \dots, S\}$, we query the source and reference clouds to form separate neighborhoods
\(\mathcal N_{P^{\rm S}}^{(s)}\), \(\mathcal N_{P^{\rm R}}^{(s)}\) and their union
\(\mathcal N_{\text{joint}}^{(s)}=\mathcal N_{P^{\rm S}}^{(s)}\cup \mathcal N_{P^{\rm R}}^{(s)}\).
On these sets we compute the same family of features as in the single scale case.
The overall feature vector for this anchor point is denoted by 
$\mathbf{f} \in \mathbb{R}^{5S + 3}$.
It contains five per-scale features
$\{H(\mathcal{N}_{\mathrm{joint}}^{(s)}), H(\mathcal{N}_{\mathrm{sep}}^{(s)}), D_\lambda^{(s)}, \rho_{\mathrm{joint}}^{(s)}, \rho_{\mathrm{sep}}^{(s)}\}$,
and three global features $\{c,d,b\}$. 



\noindent \textbf{Multiscale Attention.}
To further alleviate the trade-off of neighborhood scale, we use multiscale attention learned from the feature vector to weight the scale-dependent features.
Specifically, we split the
$(5S+3)$-dimensional feature vector $\mathbf{f}$ into $S$
8-dimensional parts, each containing 5 scale-specific features of a given scale and the 3 global features:
\vspace{-.2cm}
\begin{multline}
\mathbf{k}^{(s)}= \mathbf{v}^{(s)} = \\
\big[H(\mathcal{N}_{\mathrm{joint}}^{(s)}),\,
H(\mathcal{N}_{\mathrm{sep}}^{(s)}),\,
D_\lambda^{(s)},\,
\rho_{\mathrm{joint}}^{(s)},\,
\rho_{\mathrm{sep}}^{(s)},\,
c,d,b\big].
\end{multline}
A query-learning multilayer perceptron (MLP) \(\phi:\mathbb{R}^{5S+3}\!\to\!\mathbb{R}^{8}\) produces the query from the feature vector as
\vspace{-2mm}
\begin{equation}
\mathbf{q}=\phi(\mathbf{f})\in\mathbb{R}^{8}.
\label{eq:q_input}
\end{equation}

We then use multihead attention, where per-head tempered softmax values for scale $s$ are computed as
\vspace{-2mm}
\begin{equation}
\alpha_{i}^{(s)} \;=\; \mathrm{Softmax}\bigg(\frac{\big(\mathbf{W}^{\rm q}_i\mathbf{q}\big)\,\big(\mathbf{W}^{\rm k}_i\mathbf{k}^{(s)}\big)^{\!\top}}{\sqrt{2}\,\tau}\bigg), i\in \{ 1,2,3,4 \},
\label{eq:scale_attn}
\end{equation}
where $i$ is the head index, $\mathbf{W}^{\rm q}_i, \mathbf{W}^{\rm k}_i \in \mathbb{R}^{2 \times 8}$ are projection matrices, and \(\tau>0\) is a temperature that controls the sharpness of scale selection.
Each head aggregates values over the three scales using a value projection 
$\mathbf{W}^{\rm v}_{i} \in \mathbb{R}^{2 \times 8}$ as
\vspace{-2mm}
\begin{equation}
\tilde{\mathbf{v}}_{i} \;=\; \sum_{s=1}^{3}\alpha_{i}^{(s)}\,\big(\mathbf{W}^{\rm v}_{i}\mathbf{v}^{(s)}\big), \quad i\in \{ 1,2,3,4 \}.
\vspace{-1mm}
\end{equation}
We concatenate the outputs from all heads, project it by 
$\mathbf{W}^{\rm o} \in \mathbb{R}^{8 \times 8}$ and perform layer normalization to obtain
\begin{equation}
\tilde{\mathbf{f}} \;=\; \mathrm{LayerNorm}\!\left(\mathbf{W}^{\rm o}\,[\tilde{\mathbf{v}}_{1}^{\top},\dots,\tilde{\mathbf{v}}_{4}^{\top}]^{\top}\right)\in\mathbb{R}^{8}.
\end{equation}
We pass $\tilde{\mathbf{f}}$ through an encoder, where we use PointTransformer~\cite{zhao2021point}, followed by a three-layer MLP with ReLU activations, to predict the alignment error $E_{\text{align}}$.
Compared with plain feature concatenation, the attention mechanism mitigates multicollinearity among multiscale features and learns point-wise scale preferences.

\vspace{-3mm}

\section{Evaluation}
\label{sec:evaluate}

To approximate real-world deployment, we construct realistic datasets and compare against state-of-the-art misalignment classifiers. For the regression task, we adapt these classifiers by replacing their classification heads with a scalar predictor and training them end-to-end with an $\ell_2$ loss on our labels. The training details are provided in Appendix~\ref{appendix:traininssetting}.

\vspace{-3mm}

\subsection{Datasets for Evaluation}
\label{sec:eval_datasets}

\noindent \textbf{nuScenes–ICP.}
From nuScenes~\cite{nuplan}, we extract \num{30000} adjacent-frame pairs downsampled by 0.5m voxel size, producing high spatial overlap. Each pair is registered with ICP~\cite{132043}.
The dataset is divided by scene ID: scenes 0–700 for training, 701–750 for validation, and 751–850 for testing.

\noindent \textbf{nuScenes–ICP (noisy).}  
To simulate global navigation satellite system~(GNSS)-degraded urban segments, we perturb the source point cloud with Gaussian noise before registration by applying a random rigid perturbation with translation noise standard deviation $[2.0,2.0,0.2]$\,m (X/Y/Z) and rotation noise standard deviation $[10^\circ,2^\circ,2^\circ]$ (yaw/roll/pitch). Such perturbation levels are chosen because local registration methods like ICP are sensitive to the initial pose accuracy and are only reliable when the initial misalignment is within roughly 2\,m and $10^\circ$~\cite{lim2023quatro-plusplus}. The other settings remain as in nuScnese-ICP.

\noindent \textbf{KITTI–GeoTransformer.}
From KITTI odometry~\cite{Geiger2012CVPR}, we form \num{23000} pairs with a random interframe gap uniformly sampled from 2 to 20 frames and downsampled by 0.5\;m voxel size. This simulates registration scenarios with low overlap and scan dropout. As plain ICP is unreliable without a good initialization under low overlap, we register these pairs using GeoTransformer~\cite{qin2022geometric}.
We use sequences 08–09 for training/validation and sequence 10 for testing.


\vspace{-2mm}

\subsection{Evaluation Results}
We train and evaluate on the three datasets described above and report the root mean square error (RMSE), the mean absolute error (MAE), and the coefficient of determination (\(R^2\)). 
For our method, we consider three radii \(r_1=7.5\) \(\mathrm{m},\) 
\(r_2=4.0\,\mathrm{m},\) 
\(r_3=2.5\,\mathrm{m}\), and use temperature \(\tau=0.6\). We report the results of our method with different \(\tau\) in Appendix.~\ref{appendix:temp}.

For all baselines, we follow the same settings as reported in the original papers. e.g., FACT and CorAl both use a single adaptive radius within $[0.5, 7.5]$\,m.
Results in Table~\ref{tab:all-in-one} show that our method achieves the best performance on all datasets in terms of all three metrics RMSE, MAE, and $R^2$.
Notably, on nuScenes-ICP, most methods have \(\mathrm{RMSE}\) slightly above \(\mathrm{MAE}\), indicating few large outliers. In contrast, on KITTI--GeoTransformer, \(\mathrm{RMSE}\) is markedly larger than \(\mathrm{MAE}\), indicating heavier-tailed errors that are harder to detect.

\begin{figure}
    \centering
    \vspace{-10mm}
    \begin{subfigure}[t]{\linewidth}
        \centering
        \includegraphics[width=0.9\linewidth]{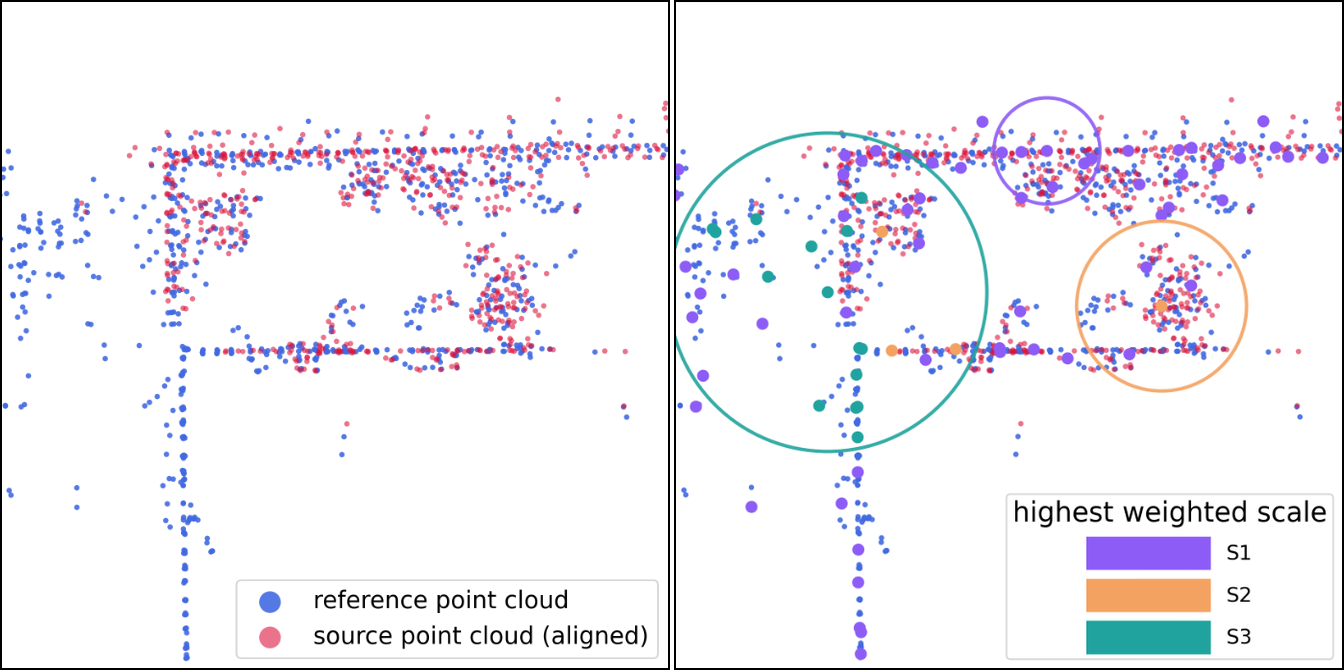}
        \caption{The case of poor registration with $E_{\text{align}} = 0.08$\,m on KITTI. The bottom-left area features a non-overlapping region. Colored circles visualize the neighborhood ranges of the corresponding scales.}
        \label{fig:pc1}
    \end{subfigure}

    \vspace{0.5em}

    \begin{subfigure}[t]{\linewidth}
        \centering
        \includegraphics[width=0.9\linewidth]{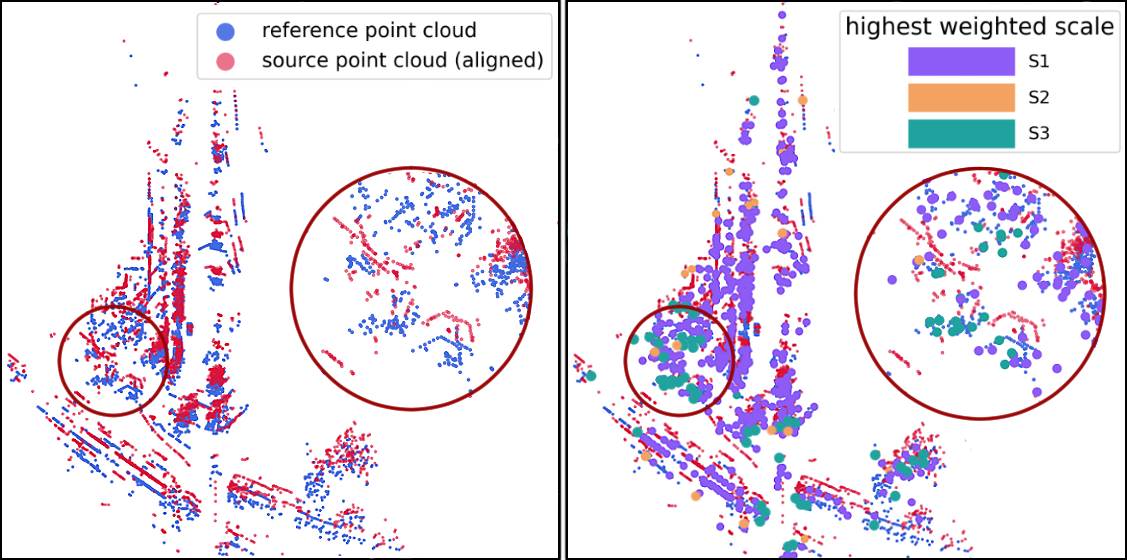}
        \caption{The case of poor registration with $E_{\text{align}} = 2.46$\,m on nuScenes. The red circle zooms in on a challenging region where larger neighborhoods are preferred.}
        \label{fig:pc2}
    \end{subfigure}
    \vspace{-0.5em}
    \caption{Visualization of the registered point clouds. \textbf{Left:} reference (blue) vs. estimated-aligned source (red). \textbf{Right:} scale preference on anchor points.}
    \label{fig:pc_all}
\end{figure}

To better understand the cause of the improvement, we visualize the registered point clouds in Fig.~\ref{fig:pc_all}, and the per-anchor-point scale that is most highly weighted by MATTER's attention. MATTER adaptively favors different radii, while FACT sets its single radius to $7.5$\;m for all anchor points in~Fig.~\ref{fig:pc_all}(\subref{fig:pc1}) and $95\%$ of the anchor points in~Fig.~\ref{fig:pc_all}(\subref{fig:pc2}). Specifically, our attention mechanism favors the smallest scale S1 in dense, well-overlapping regions where the local geometry better approximates a 3D Gaussian, making the closed-form differential-entropy descriptors more reliable. When correspondences are absent or alignment is poor, the model adaptively enlarges the radius. In Fig.~\ref{fig:pc_all}(\subref{fig:pc1}), the left side of the reference point cloud lacks corresponding points from the source, which triggers the largest scale S3 so that anchors still cover areas that contain matches. In Fig.~\ref{fig:pc_all}(\subref{fig:pc2}), the largest scale S3 is favored in the red circles because small neighborhoods fail to capture counterparts of both clouds. The midscale S2 is generally less preferred but appears in cluttered medium-thickness structures. 

\begin{table*}[t]
\centering
\small 
\begin{tabular}{l ccc ccc ccc}
\toprule
Methods & \multicolumn{3}{c}{nuScenes} & \multicolumn{3}{c}{nuScenes (noisy)} & \multicolumn{3}{c}{KITTI} \\
\cmidrule(lr){2-4}\cmidrule(lr){5-7}\cmidrule(lr){8-10}
& RMSE (m)$\downarrow\!$ & MAE (m)$\downarrow\!$ & $R^2$$\uparrow\!$ & RMSE (m)$\downarrow\!$ & MAE (m)$\downarrow\!$ & $R^2$$\uparrow\!$ & RMSE (m)$\downarrow\!$ & MAE (m)$\downarrow\!$ & $R^2$$\uparrow\!$ \\
\midrule
NDT Score$\ast\dag$~\cite{almqvist2018learning}        & 0.568 & 0.268 & 0.956 & 0.602 & 0.307 & 0.953 & 0.623 & 0.329 & 0.950 \\
CorAl$\ast$~\cite{adolfsson2021coral}                   & 0.576 & 0.311 & 0.946 & 0.603 & 0.321 & 0.953 & 0.622 & 0.129 & 0.953 \\
KPConv-based$\ast\dag$~\cite{camous2022deep}$\!\!\!$          & 0.358 & 0.152 & 0.972 & 0.501 & 0.227 & 0.956 & 0.551 & 0.258 & 0.917 \\
FACT$\ast$~\cite{dillen2025fact}                        & 0.323 & 0.162 & 0.983 & 0.466 & 0.239 & 0.972 & 0.491 & 0.112 & 0.969 \\
MATTER (Ours)                                           & \textbf{0.243} & \textbf{0.147} & \textbf{0.990} &
\textbf{0.415} & \textbf{0.224} & \textbf{0.978} &
\textbf{0.336} & \textbf{0.095} & \textbf{0.985} \\
\bottomrule
\end{tabular}
\vspace{-.2cm}
\caption{Performance comparison on nuScenes and KITTI. Best scores in each column are in bold. Methods marked with $\ast$ have the classification head replaced by a regression head to predict \eqref{eq:align_error}. Methods marked with $\dag$ are re-implemented due to absence of open source code. Arrows indicate the better direction (\(\downarrow\): lower is better; \(\uparrow\): higher is better). 
}
\label{tab:all-in-one}
\vspace{-4mm}
\end{table*}

\begin{figure}
  \centering
  \vspace{-20mm}
  \includegraphics[width=1.0\linewidth]{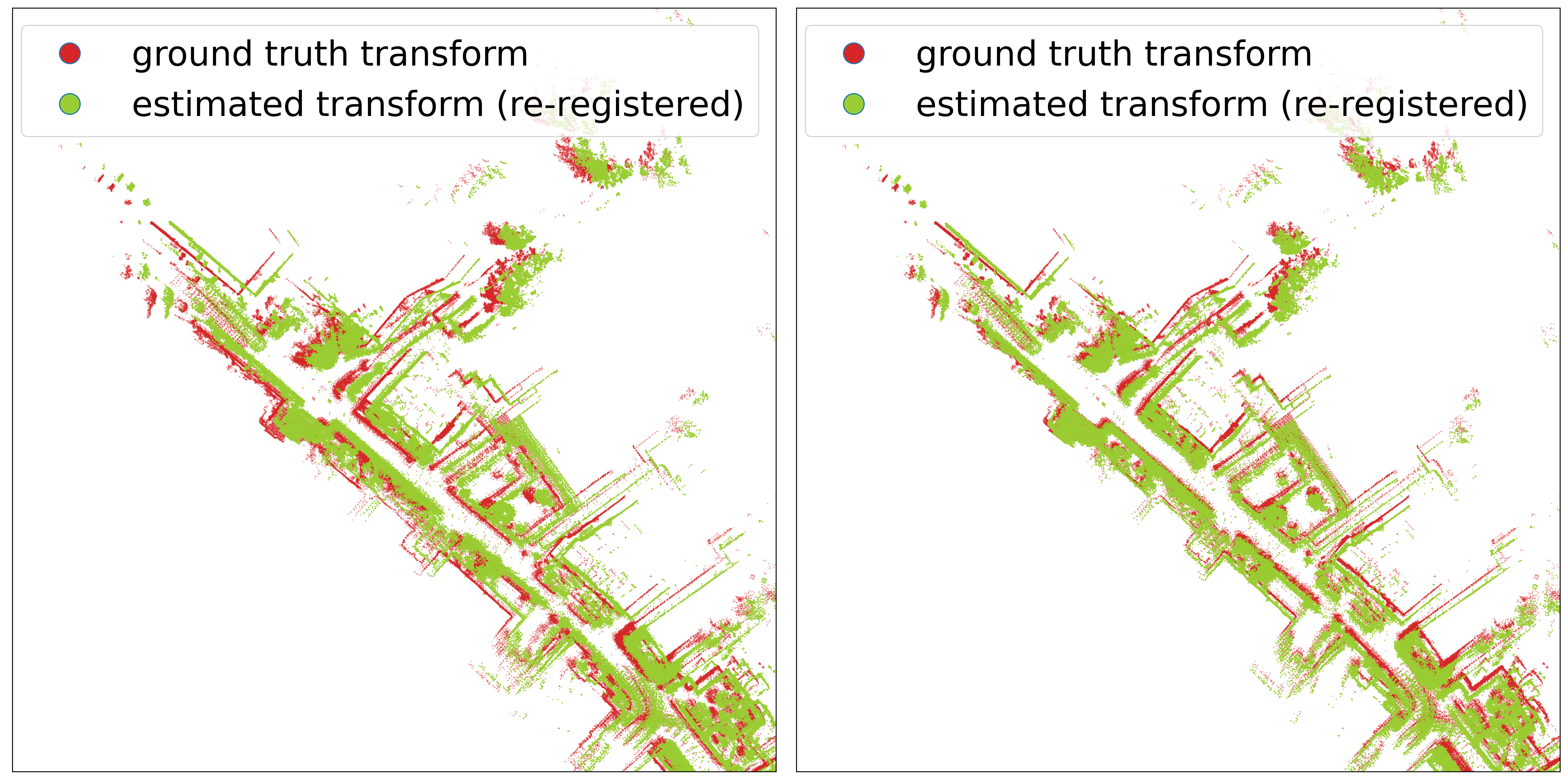}
  \caption{Visualization of re-registered point cloud sequences. \textbf{Left:} FACT, $E_{\text{align}} = 21.46$~m; \textbf{Right:} MATTER, $E_{\text{align}} = 18.35$~m.}
  \vspace{-10mm}
  \label{fig:downstream_vis}
\end{figure}



We report an ablation on the radius of the neighborhood and multiscale attention, and summarize the results in
Table~\ref{tab:ablation_radius}.
Single-scale models prefer different radii per dataset: 7.5\,m in nuScenes (high overlap) and 2.5\,m in KITTI (low/moderate overlap), revealing the limitation of a single scale.
Plain multiscale concatenation already improves over single-scale on nuScenes (noisy) and KITTI, but our point-wise scale-selector attention achieves the lowest RMSE on all datasets with only approximately 0.06\% more parameters.







\begin{table}[t]
\centering
\small 
\resizebox{\columnwidth}{!}{%
\begin{tabular}{lcccc}
\toprule
Radius & Params & nuScenes & nuScenes & KITTI \\
       &       &       &    (noisy)    &       \\
\midrule
Single scale (7.5 m)        &     3.166M      & \textbf{0.335} & \textbf{0.490} &  0.502\\
Single scale (4.0 m)        &     3.166M      & 0.489 & 0.586 &  0.554\\
Single scale (2.5 m)        &     3.166M      & 0.442 & 0.594 &  \textbf{0.483}\\
\midrule
Multiscale (No Attention)   & 3.166M & 0.378 & 0.457 &  0.468\\
Multiscale (Attention) & 3.168M & \textbf{0.243} & \textbf{0.415} & \textbf{0.336} \\
\bottomrule
\end{tabular}
}%
\vspace{-.2cm}
\caption{Ablation on Neighborhood Radius and Multiscale Attention across datasets (RMSE in meters).}
\label{tab:ablation_radius}
\vspace{-3.5mm}
\end{table}

\vspace{-3mm}

\subsection{Downstream Task}

We evaluate the impact of error estimation in a mapping protocol. On KITTI odometry sequence 10, we take one scan every 5 frames and build a map by chaining pairwise registrations with GeoTransformer, yielding an estimated pose for each scan. Overall mapping quality is measured as the point-wise mean distance in (\ref{eq:align_error}) between the last scan transformed by the estimated transform and by the ground-truth pose.
MATTER and FACT are then used to detect misaligned point cloud pairs.
For FACT, we re-register all pairs classified into the $[0.10, +\infty)$\;m error bin by replacing the estimated relative transform with the ground-truth transform. 
For MATTER, we re-register the point cloud pairs whose predicted error exceeds the threshold $0.1212$ m, resulting in the same number of re-registered pairs (38.9\% of all pairs).
Visualized results of the last 200 pairs are shown in Fig.~\ref{fig:downstream_vis}. 
MATTER achieves an alignment error $E_{\text{align}} = 18.35$\;m for the last frame, outperforming FACT's $E_{\text{align}} = 21.46$\;m.

\begin{figure}[h]
    \vspace{-.15cm}
  \centering
  \includegraphics[width=0.8\linewidth]{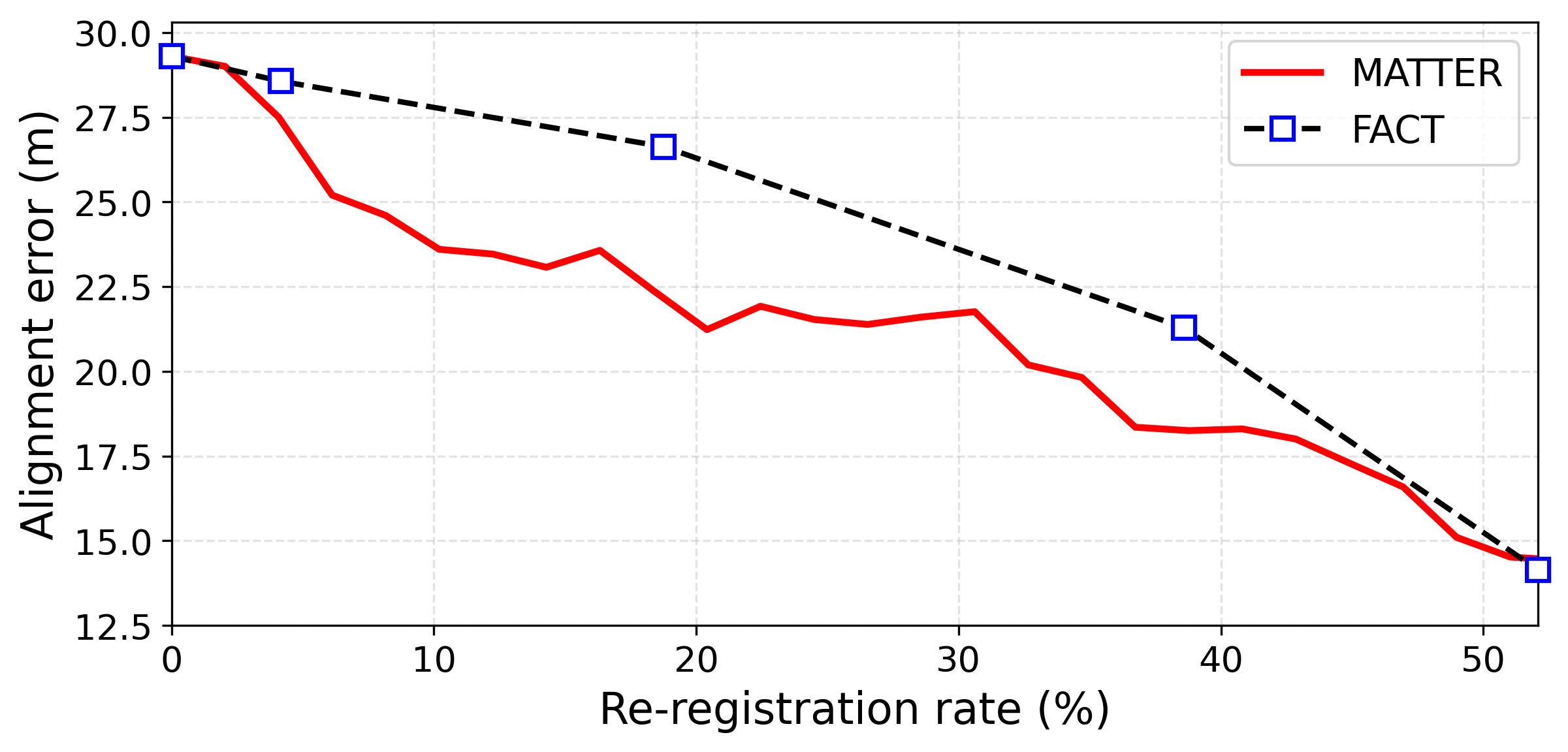}
  \vspace{-.3cm}
  \caption{The final-frame alignment error vs. re-registration rate when using MATTER/FACT for misalignment detection.}
  \vspace{-7mm}
  \label{fig:sparse}
\end{figure}

Beyond comparison with a fixed threshold, we plot in Fig.~\ref{fig:sparse} the final-frame alignment error under various re-registration rates (similar to~\cite[Fig.~5]{KRISTOFFERSSONLIND202491}). 
By correcting more frames, the final-frame alignment error keeps reducing. 
For low re-registration rates (e.g., from 7\% to 40\%), MATTER reaches significantly lower alignment error of the final frame by correcting the same amount of previous frames.

\vspace{-1mm}

\section{Conclusion}
\label{sec:conclusion}
\vspace{-2mm}


We extend point cloud misalignment assessment from classification to regression and propose MATTER, a point-wise multiscale attention framework for registration error regression. Across three datasets, MATTER achieves state-of-the-art RMSE, MAE, and R² and remains accurate under poor registration and low overlap. 
In the mapping downstream application, MATTER's error predictions result in improved mapping quality without needing more re-registrations.

\small
\bibliographystyle{IEEE}
\bibliography{sample}

\vfill\clearpage

\appendix

\section{Training Settings}
\label{appendix:traininssetting}

All models were trained on an NVIDIA A100 GPU using the Adam optimizer with a batch size of 16 for 200 epochs. The initial learning rate was set to $1\times10^{-4}$ with weight decay $1\times10^{-4}$. A step learning-rate scheduler was employed, decaying the learning rate every 10 epochs with a decay factor of 0.825. The backbone network followed the PointTransformer~\cite{zhao2021point} architecture consisting of 4 transformer--transition blocks and a transformer feature dimension of 256. During training, data augmentation included random dropout, scaling, and translation. For point sampling, farthest point sampling (FPS) was applied to obtain 2048 points per point cloud. Differential entropy estimation used a logarithmic stability parameter $\log \epsilon = -18$ with a rejection threshold of 0.20. Sinkhorn divergence was computed with $p=2$ and blur parameter 0.05.

\section{Related Works}
\label{appendix:relatedwork}

Existing approaches for post-registration quality assessment can be broadly divided into objective-function-based and learning-based methods. 
Classical registration pipelines typically rely on the final value of the objective during optimization, such as closest-point residuals~\cite{132043}, score functions~\cite{liao2020point}, or entropy gain~\cite{tustison2010point}, as an indicator of alignment quality. 
However, these objectives are designed to measure fitness during optimization rather than to measure the true pose error, and their values are highly sensitive to local minima, noise, and scene geometry. 
As a result, they do not exhibit a reliable or monotonic relationship with the actual alignment error~\cite{access_loss}. 
As noted in~\cite{adolfsson2021coral}, such metrics ``provide limited information on whether the point clouds are correctly aligned once registration has been carried out,'' meaning that a low objective value does not necessarily imply correct alignment.

To overcome this limitation, several works have proposed learning-based point cloud misalignment classification (PCMC) methods, which predict the registration quality directly from geometric features. 
Early approaches~\cite{yin2019failure, almqvist2018learning} employ manually designed features combined with logistic regression to perform binary classification of alignment quality. 
CorAl~\cite{adolfsson2021coral} improves feature representation by fitting a local 3D Gaussian distribution to each point neighborhood and using the resulting differential entropy as a descriptor. 
More recently, FACT~\cite{dillen2025fact} extends the formulation from binary to multi-class classification, enabling a finer categorization of misalignment levels and representing the current state of the art in PCMC.

Despite these advances, existing PCMC methods remain limited to discrete classification, providing only coarse alignment quality labels rather than continuous estimates of the true registration error. 
This motivates the need for point cloud misalignment regression, which directly predicts the alignment error in a continuous domain.

\section{Feature Extraction}
\label{appendix:features}

For each pair $(P^\mathrm{S}, P^\mathrm{R})$, we first transform both point clouds into a common coordinate frame and apply farthest point sampling (FPS)~\cite{fps} to each of them to select the same number of anchor points from $P^\mathrm{S}$ and $P^\mathrm{R}$, respectively.

\medskip
\noindent\textbf{Differential Entropy.}
Following CorAl~\cite{adolfsson2021coral}, we use differential entropy as features. Let $\mathcal{N}_{P^{\rm S}}$ and $\mathcal{N}_{P^{\rm R}}$ denote the points from the source and reference scans within the neighborhood of anchor point $p$, respectively. Let the joint neighborhood be $\mathcal{N}_{\mathrm{joint}}=\mathcal{N}_{P^{\rm S}}\cup\mathcal{N}_{P^{\rm R}}$. We also use $\mathcal{N}_\mathrm{sep}$ to refer to $\mathcal{N}_{P^\mathrm{S}}$ if the anchor point belong to $P^\mathrm{S}$, and to $\mathcal{N}_{P^\mathrm{R}}$ if it belongs to $P^\mathrm{R}$. Assuming that each neighborhood has an approximately isotropic Gaussian distribution, its differential entropy has a closed form.
\begin{equation}
H(\mathcal{N}) \;=\; \tfrac{1}{2}\,\ln\!\big[(2\pi e)^3\det(\Sigma)\big],
\label{eq:entropy}
\end{equation}
where $\Sigma$ is the sample covariance of the points in neighborhood $\mathcal{N}$. Well-aligned point cloud pairs yield small gaps between $H(\mathcal{N}_{\mathrm{joint}})$ and $H(\mathcal{N}_{\mathrm{sep}})$, whereas misalignment inflates dispersion when the two sets are merged, increasing the gaps.

\medskip
\noindent\textbf{Sinkhorn Divergence.}
To directly measure the discrepancy between the distributions of points in the aligned source vs. reference neighborhoods, we employ the Sinkhorn divergence as in \cite{dillen2025fact}. We treat the sets $\mathcal{N}_{P^{S}}$ and $\mathcal{N}_{P^{R}}$ as two discrete distributions of unit mass. Let $C_{ij} = \|x_i - y_j\|^2$ be the transport cost between a point $x_i \in \mathcal{N}_{P^{S}}$ and a point $y_j \in \mathcal{N}_{P^{R}}$. The entropic optimal transport distance between the two point sets is defined as:
    {\small
    \begin{equation}
    W_\lambda(\mathcal{N}_{P^{S}}, \mathcal{N}_{P^{R}})=
    \min_{\pi \in \Pi(\mathcal{N}_{P^{S}}, \mathcal{N}_{P^{R}})}
    \sum_{i,j} C_{ij}\,\pi_{ij} + \lambda \sum_{i,j} \pi_{ij} \ln \pi_{ij}~,
    \label{eq:entropic_ot}
    \end{equation}
    }
    where $\Pi(\mathcal{N}_{P^{S}}, \mathcal{N}_{P^{R}})$ is the set of all transport plans (coupling matrices) that respect the marginal distributions of $\mathcal{N}_{P^{S}}$ and $\mathcal{N}_{P^{R}}$, and $\lambda>0$ is a regularization coefficient. We therefore use the Sinkhorn divergence:
    {\small
    \begin{align}
    D_\lambda(\mathcal{N}_{P^{S}}, \mathcal{N}_{P^{R}}) 
       &= W_\lambda(\mathcal{N}_{P^{S}}, \mathcal{N}_{P^{R}}) \notag \\
       &\quad - \tfrac{1}{2}\Big[W_\lambda(\mathcal{N}_{P^{S}}, \mathcal{N}_{P^{S}})
                     + W_\lambda(\mathcal{N}_{P^{R}}, \mathcal{N}_{P^{R}})\Big]~,
    \label{eq:sinkhorn_div}
    \end{align}
    }
    which is non-negative, symmetric, and equals zero if and only if the two point distributions coincide exactly. 

    \medskip
\noindent \textbf{Reliability Features.}
Inspired by \cite{dillen2025fact}, for each anchor point, we attach features indicating neighborhood trustworthiness: coverage ratios $\rho_{\mathrm{joint}}=\big|\mathcal{N}_{\mathrm{joint}}\big|/\big|P_{\mathrm{joint}}\big|$ and
$\rho_{\mathrm{sep}}=\big|\mathcal{N}_{\mathrm{sep}}\big|/\big|P_{\mathrm{sep}}\big|$ (with $\big|P_{\mathrm{joint}}\big|=\big|P^{\rm S}\big|+\big|P^{\rm R}\big|$, $\big|P_{\mathrm{sep}}\big|= \big|P^{S}\big|$ or $\big|P^{S}\big|$, given by the point cloud that anchor point $p$ belongs to.); a co-visibility score $c\!\in[0,1]$ computed with a visibility operator~\cite{visibility}; the distance $d$ to the LiDAR sensor; and a binary source flag $b\!\in \{0,1\}$ designating which point cloud a point originates. Within the attention mechanism they down-weight anchors that are non-covisible, far/sparse, and up-weighting well-observed regions.

\section{Ablation on Temperature}
\label{appendix:temp}

Temperature \(\tau\) is a key hyperparameter that controls the smoothness of multiscale fusion: a smaller \(\tau\) yields sharper point-wise scale choices, while a larger \(\tau\) averages across scales. We report the results of our method with different \(\tau\), as shown in Table~\ref{tab:temp_ablate}.
Among the considered values, the optimal are \(\tau=0.6\) on nuScenes, \(\tau=0.8\) on nuScenes (noisy), and \(\tau=0.4\) on KITTI, indicating that sharper selection benefits low-overlap regimes, while slightly smoother fusion helps under noisy initialization.

\clearpage
\section{Larger Visualizations}
\label{appendix:largevis}
\FloatBarrier

\begin{table}[t]
\centering
\small
\begin{tabular}{lccc}
\toprule
\(\tau\) & nuScenes & nuScenes (noisy) & KITTI \\
\midrule
1.0                    & 0.382 & 0.456 &  0.451\\
0.8                    & 0.272 & \textbf{0.407} &  0.365\\
0.6                    & \textbf{0.243} & 0.415 &  0.336\\
0.4                    & 0.295 & 0.418 &  \textbf{0.305}\\
\bottomrule
\end{tabular}
\caption{Ablation on temperature \(\tau\) for the scale-selector attention (RMSE in meters).}
\label{tab:temp_ablate}
\end{table}

\begin{figure*}[t]
    \centering
    \begin{subfigure}[t]{0.49\textwidth}
        \centering
        \includegraphics[width=\linewidth]{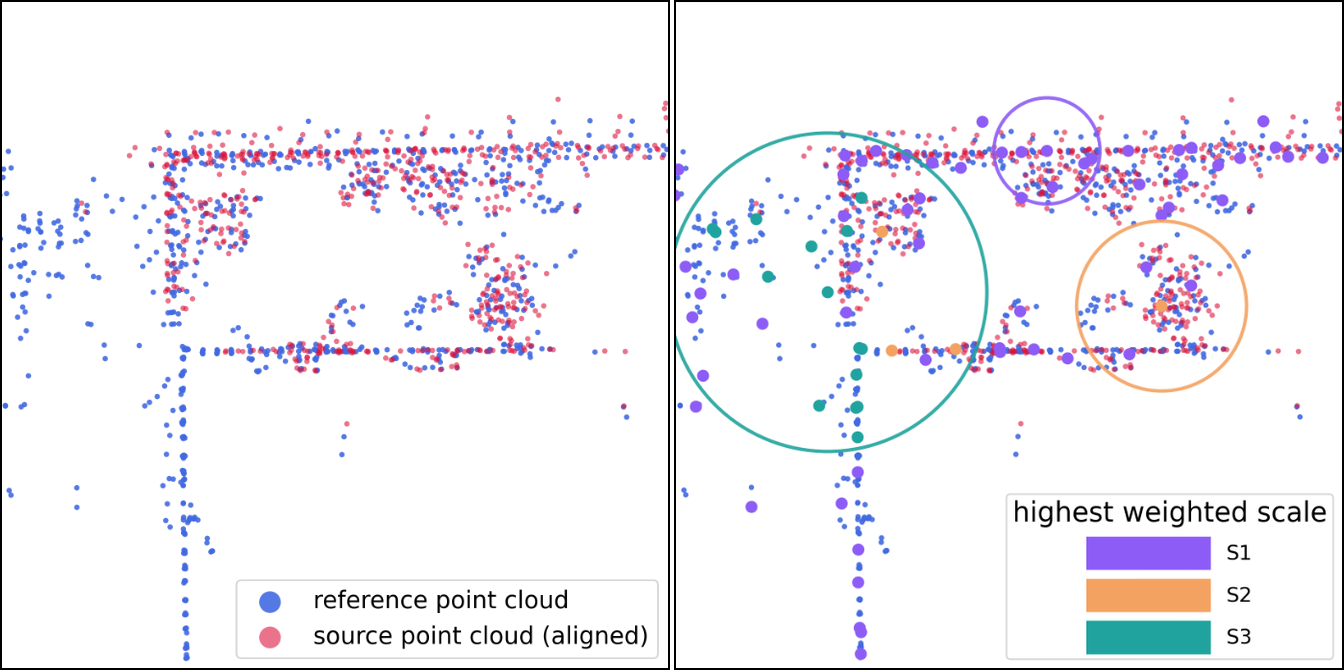}
        \caption{Poor registration on KITTI with $E_{\text{align}} = 0.08$\,m. The bottom-left area is non-overlapping. Colored circles visualize neighborhood ranges of the corresponding scales.}
        \label{fig:pc1_appendix}
    \end{subfigure}
    \hfill
    \begin{subfigure}[t]{0.49\textwidth}
        \centering
        \includegraphics[width=\linewidth]{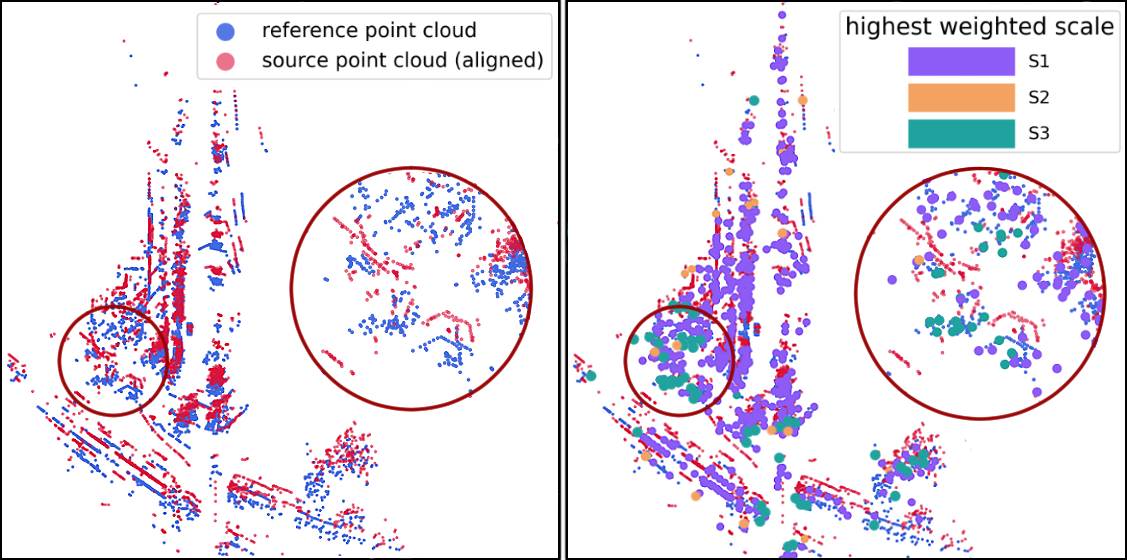}
        \caption{Poor registration on nuScenes with $E_{\text{align}} = 2.46$\,m. The red circle highlights a challenging region where larger neighborhoods are preferred.}
        \label{fig:pc2_appendix}
    \end{subfigure}

    \vspace{0.6em}

    \includegraphics[width=0.90\textwidth]{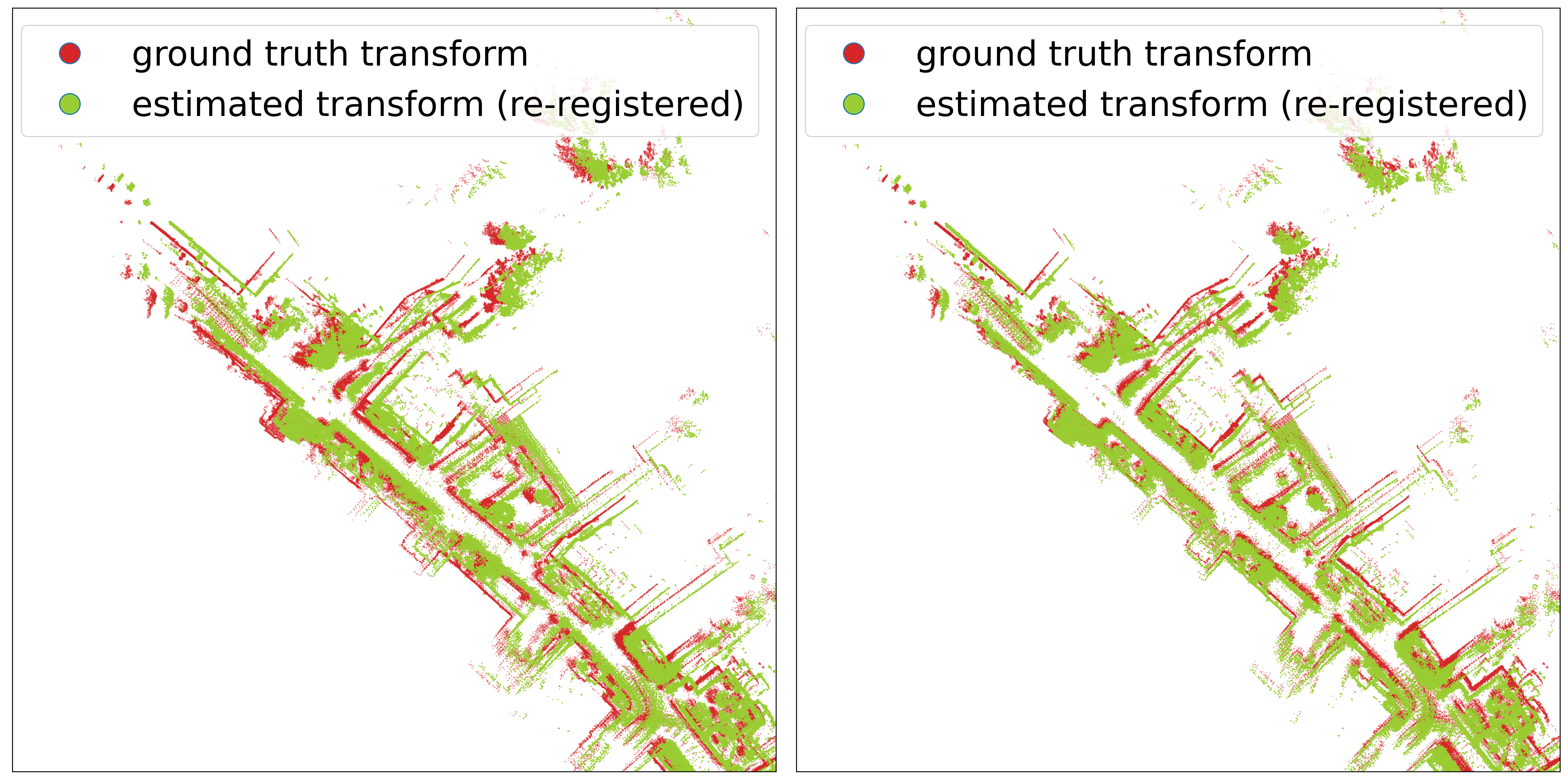}
    \caption{Larger visualizations from the main paper. Top: registered point clouds and scale preference. Bottom: re-registered point cloud sequences (\textbf{Left:} FACT, $E_{\text{align}} = 21.46$~m; \textbf{Right:} MATTER, $E_{\text{align}} = 18.35$~m).}
    \label{fig:largevis_all}
\end{figure*}

For improved readability and interpretation, we include larger versions of Fig.~2 and Fig.~3 from the main paper in  Fig.~\ref{fig:largevis_all}.

\end{document}